\pdfoutput=1

\documentclass[11pt]{article}

\usepackage[final]{ACL2023}

\usepackage{times}
\usepackage{latexsym}

\usepackage{amsmath,graphicx}
\usepackage{adjustbox}
\usepackage{colortbl}
\usepackage{appendix}
\usepackage{color}
\usepackage{tabularray}
\UseTblrLibrary{diagbox}

\usepackage[T1]{fontenc}

\usepackage[utf8]{inputenc}

\usepackage{microtype}

\usepackage{inconsolata}

\title{Exploring Robustness in Doctor-Patient Conversation Summarization: An Analysis of Out-of-Domain SOAP Notes}

\author{Yu-Wen Chen, Julia Hirschberg\\
 Department of Computer Science, Columbia University, United States \\
 \texttt{yu-wen.chen@columbia.edu, julia@cs.columbia.edu} 
 } 

\begin{document}
\maketitle
\begin{abstract}
Summarizing medical conversations poses unique challenges due to the specialized domain and the difficulty of collecting in-domain training data. In this study, we investigate the performance of state-of-the-art doctor-patient conversation generative summarization models on the out-of-domain data. We divide the summarization model of doctor-patient conversation into two configurations: (1) a general model, without specifying subjective (S), objective (O), and assessment (A) and plan (P) notes; (2) a SOAP-oriented model that generates a summary with SOAP sections. We analyzed the limitations and strengths of the fine-tuning language model-based methods and GPTs on both configurations. We also conducted a Linguistic Inquiry and Word Count analysis to compare the SOAP notes from different datasets. The results exhibit a strong correlation for reference notes across different datasets, indicating that format mismatch (i.e., discrepancies in word distribution) is not the main cause of performance decline on out-of-domain data. Lastly, a detailed analysis of SOAP notes is included to provide insights into missing information and hallucinations introduced by the models.

\end{abstract}

\section{Introduction}

Automatically generated summary notes of doctor-patient conversations could improve the healthcare system. First, the generated notes serve as a valuable resource, allowing doctors to review and validate the information from the conversation with a patient, ensuring that vital information is noticed. In addition, the summary notes can be integrated into hospitalization risk prediction models~\cite{song2022clinical}, empowering healthcare professionals with data-driven insights to make more precise clinical decisions.

However, summarizing doctor-patient conversations poses distinct challenges owing to its specialized domain. Specifically, medical conversations often involve highly specialized terminology that requires domain-specific knowledge to understand and summarize accurately. In addition, it is preferable to structure the generated note with  \textbf{S}ubjective (information reported by the patient), \textbf{O}bjective (objective observations), \textbf{A}ssessment (doctor's evaluation), and \textbf{P}lan (future care plan) (SOAP). SOAP format is preferable because it is widely utilized by healthcare providers to document a patient's progress, providing an organized framework that reduces communication confusion among healthcare professionals. These challenges hinder the direct application of general-purpose summarization techniques to doctor-patient conversations, underscoring the need for a specialized model.

Doctor-patient conversation summarization has attracted significant attention recently~\cite{joshi2020dr, krishna2020generating, zhang2021leveraging,grambow2022domain, abacha2023overview}. In 2023, the MEDIQA-Chat Challenge~\cite{abacha2023overview} attracted 120 registered teams from the academy and industry. Although various methods are proposed in MEDIQA-Chat, it remains a challenging field that needs further investigation. First, MEDIQA-Chat focuses on in-domain training and testing. However, cross-dataset analysis for doctor-patient conversation summarization is crucial because collecting in-domain training data is usually challenging given the constraints imposed by privacy and security concerns. Second, a detailed assessment of performance across SOAP note categories is essential. Such insights into the performance of each category can play a pivotal role in developing improved model structures and designing more effective evaluation metrics.

In this study, we investigate cross-dataset performance of state-of-the-art (SOTA) doctor-patient summarization models. Our focus is on generative summarization models because the real-world clinical notes are in an abstractive format. The experiments were conducted on English datasets as the setting of most previous studies. The results of SOAP notes are evaluated separately to gain a deeper understanding of the strengths and limitations of the current models. We hope our result can offer new insights for future research in developing a robust doctor-patient summarization model for real-world scenarios.

\section{Related Work}

The MEDIQA-Chat challenge~\cite{abacha2023overview} separated doctor-patient conversation summarization into different tasks. Models designed for Task A predict the topic category of the conversation and then generate notes. The Task A models are closer to a general-purpose summarization model, producing notes without specifying distinct sections. In the top performance models, Wanglab~\cite{ giorgi2023wanglab} fine-tuned a FLAN-T5 model~\cite{chung2022scaling} for summarization and note classification. SummQA~\cite{mathur2023summqa} used BioBERT~\cite{lee2020biobert} to support the section classification, MiniLM~\cite{wang2020minilm} to select the prompt for GPT4, and GPT4 to predict the section class and generated the final note. The Cadence~\cite{sharma2023team} model fine-tuned BART-large on the SAMSum dataset, followed by fine-tuning on the augmented dataset. In addition, a N-pass summarization was employed to handle long conversations. 

Models designed for Task B are SOAP-oriented, generating notes with SOAP sections. In the top performance models, WangLab used instructor~\cite{su2022one} to select the top-k conversation that is similar to the testing data, then used the selected conversations and notes as the in-context learning examples for GPT4. They also achieved top performance with the fine-tuned Longformer Encoder-Decoder (LED)~\cite{beltagy2020longformer}. SummQA~\cite{mathur2023summqa} used the MiniLM~\cite{wang2020minilm} to select the prompt for the GPT4 in-context learning examples as their model for task A. GersteinLab~\cite{tang2023gersteinlab} used GPT-4 with specifically designed instruction. 

Task A in the MEDIQA-Chat challenge was evaluated on the MTS-Dialog dataset~\cite{abacha2023empirical}, which has a relatively shorter conversation and reference notes related to a specific category. Task B was focused on the ACI-BENCH~\cite{yim2023aci} dataset, which has a relatively longer conversation and a long note with SOAP sections. Most top-performance teams in Task A used fine-tuning language model (LM)-based methods, while most top-performance teams in Task B introduced GPT-based approaches. The results seem to indicate that the fine-tuning LLM-based method is more suitable for \emph{short} dialogues with a specific category of information. In contrast, the GPT-based method is preferable for the \emph{long} dialogue with detailed SOAP information~\cite{abacha2023overview}. However, in real-world scenarios, conversations may vary in length and encompass one or multiple categories of information. Therefore, in this study, we aim to understand how these models perform in an cross-dataset settings and identify potential errors made by the models.

\section{Data}\label{sec:data}
\begin{figure}[htbp!]
\centering
\includegraphics[scale=0.85]{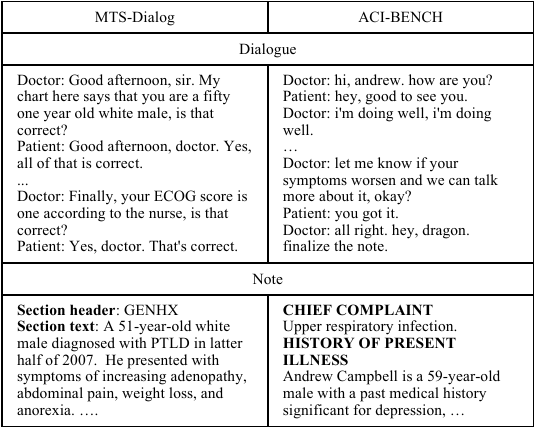}
\caption{Dataset examples. Samples in the MTS-Dialog dataset have a section header that indicates the category of the annotation and the section text, which is the main content of the notes. The samples in the ACI-BENCH dataset have one full note, where each section is separated by bold title text.}
\label{fig:dataset_examples}
\end{figure}

We use two open-source doctor-patient conversation datasets, MTS-Dialog~\cite{abacha2023empirical} and ACI-BENCH~\cite{yim2023aci}. Both datasets contain doctor-patient conversations, the corresponding note of the conversation, and the category of the note. Figure~\ref{fig:dataset_examples} illustrates the samples in the two datasets, and Table~\ref{tab:statistic} summarizes the dataset statistics. The number of tokens is calculated using the \emph{google/flan-t5-large} tokenizer\footnote{\url{https://huggingface.co/google/flan-t5-large}}. 

Compared with the two datasets, the MTS-Dialog dataset contains relatively shorter conversation, and the reference note follows a concise format, comprising either a few words or a one-paragraph structure with a section header specifying the note category. In contrast, the conversations in the ACI-BENCH dataset are relatively longer, and the reference notes includes all SOAP sections. 

\begin{table}[htpb!]
\centering
\resizebox{\columnwidth}{!}{%
\begin{tabular}{|cccc|}
\hline
\multicolumn{1}{|c|}{}          & \multicolumn{1}{c|}{Train} & \multicolumn{1}{c|}{Valid} & Test \\ \hline
\multicolumn{4}{|c|}{Number of samples}                                                          \\ \hline
\multicolumn{1}{|c|}{MTS-Dialog} & \multicolumn{1}{c|}{1,201}        & \multicolumn{1}{c|}{100}          & 200         \\ \hline
\multicolumn{1}{|c|}{ACI-BENCH} & \multicolumn{1}{c|}{67}    & \multicolumn{1}{c|}{20}    & 40   \\ \hline
\multicolumn{4}{|c|}{Number of tokens of dialogue (mean/max)}                                    \\ \hline
\multicolumn{1}{|c|}{MTS-Dialog} & \multicolumn{1}{c|}{152.4 / 2343}   & \multicolumn{1}{c|}{129.27 / 820}   & 144.2 / 793   \\ \hline
\multicolumn{1}{|c|}{ACI-BENCH}  & \multicolumn{1}{c|}{1931.49 / 4642} & \multicolumn{1}{c|}{1814.95 / 2608} & 1824.4 / 3560 \\ \hline
\multicolumn{4}{|c|}{Number of tokens of note (mean/max)}                                        \\ \hline
\multicolumn{1}{|c|}{MTS-Dialog} & \multicolumn{1}{c|}{59.63 / 1580}   & \multicolumn{1}{c|}{53.9 / 406}     & 57.4 / 530    \\ \hline
\multicolumn{1}{|c|}{ACI-BENCH}  & \multicolumn{1}{c|}{663.22 / 1388}  & \multicolumn{1}{c|}{680.3 / 1176}  & 647.7 / 1291  \\ \hline
\end{tabular}%
}
\caption{Statistic of MTS-Dialog and ACI-BENCH dataset.}
\label{tab:statistic}
\end{table}

We categorized the note in the MTS-Dialog dataset and divided the note in ACI-BENCH dataset into S, O, or AP categories for analysis. Note that we merged A and P as AP because these are merged into AP in the ACI-BENCH dataset, making it difficult to separate them into A and P. Table~\ref{tab:mapping} shows the mapping between original note categories and SOAP and the number of samples in each category. 

\begin{table}[htbp!]
\centering
\includegraphics[scale=0.85]{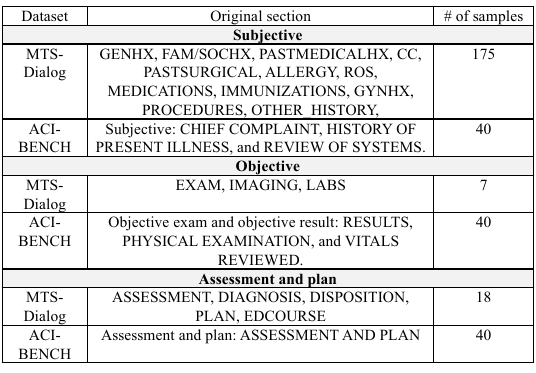}
\caption{Mapping between original note categories and SOAP.}
\label{tab:mapping}
\end{table}

\section{Methods}

\begin{figure}[htbp!]
\centering
\includegraphics[scale=0.85]{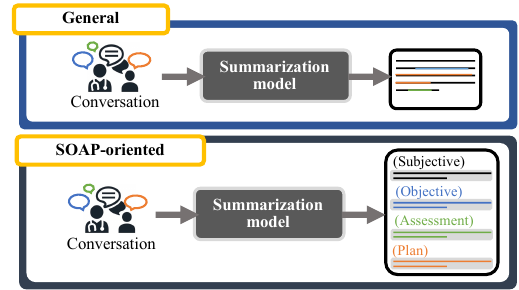}
\caption{Illustration of the general and SOAP-oriented configurations.}
\label{fig:model_definition}
\end{figure}

We divided the summarization model for doctor-patient conversation into general and SOAP-oriented configurations (illustrated in Figure~\ref{fig:model_definition}). 
In this study, we investigate the current SOTA models of each configuration in a cross-dataset setting. Our research question is:

\textbf{RQ1: How do current SOTA doctor-patient conversation summarization models perform on out-of-domain datasets, and what causes the performance decline?}

\subsection{Cross-dataset analysis of general model}

\begin{figure}[htbp!]
\centering
\includegraphics[scale=0.85]{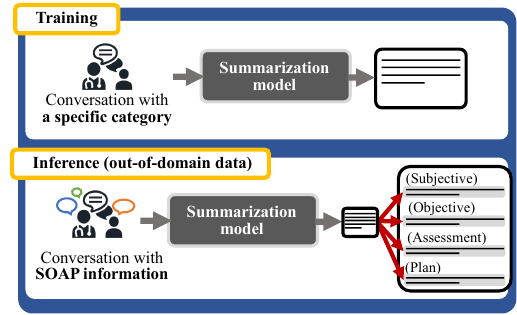}
\caption{Analysis of fine-tuning LM-based general model.}
\label{fig:general_model}
\end{figure}

We analyzed the limitations of directly applying a general configuration for doctor-patient conversation summarization. Because the model does not consider generating S, O, A, and P notes separate tasks, the model may emphasize some information more than others, thus leading to missing information issues in the generated note. Therefore, we examined the following research question:

\textbf{RQ2: What information is more likely to be missing in SOAP for model with a general configuration? (Figure~\ref{fig:general_model})} Our hypothesis is that objective information can easily be excluded from summaries. Objective information usually includes numerical information that holds significant importance in medical contexts. The number could represent the quantity of medication administered to the patient or the values derived from their health examination report, serving as indispensable metrics for assessing the patient's overall health condition. However, numerical data is often considered as detailed information and thus omitted in summaries. In addition, objective information is closely associated with technical terms, making it more challenging for the LM.

\subsubsection{Model and Data}
We used the fine-tuned Flan-T5 model~\cite{chung2022scaling}, which received the top rank in the MEDIQA-Chat challenge task A, as representative for model with general configuration. The Flan-T5 model was fine-tuned with the MTS-Dialog dataset, in which the reference notes focus only on one topic in the conversation. We also included the GPT results (gpt-3.5-turbo and gpt4) for comparison. Models with the general configuration are evaluated on the ACI-BENCH dataset. Because the conversations and reference notes in the this dataset contain all SOAP information, we can analyze what categories  of information (i.e., S, O, A, or P) are missing from the generated note.

\subsection{Cross-dataset analysis of SOAP-oriented model}

\begin{figure}[htbp!]
\centering
\includegraphics[scale=0.85]{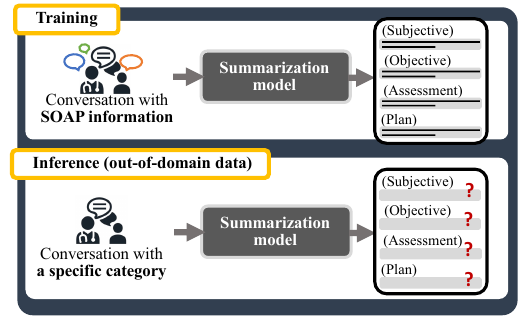}
\caption{Analysis of fine-tuning LM-based SOAP-oriented model.}
\label{fig:soap_model}
\end{figure}

The model with SOAP-oriented configuration aims to generate notes with S, O, A, and P sections. However, in real-world conditions, not all doctor-patient conversations include all of the S, O, A, and P information. For example, doctors might skip the objective information because they already have the record. They might also not mention assessments and plans because they only want to check the patient’s condition. Therefore, we ask the following research question: 

\textbf{RQ3: What SOAP-oriented model will generate if the input conversation does not include information related to a specific category? (Figure~\ref{fig:soap_model})} We hypothesize that the LM will have severe hallucination problems by generating information that does not exist in the conversation. 

\subsubsection{Model and Data}
We used the fine-tuned LED model~\cite{beltagy2020longformer}, which received top-rank performance in the MEDIQA-Chat challenge task B as representative of the SOAP-oriented model. The LED model was fine-tuned with the ACI-BENCH dataset that specifies notes into SOAP sections. We also included the GPT results for comparison. The GPT was prompted to generate a note with SOAP sections and was informed that it could skip the section if no relevant information was provided in the conversation. We evaluated the models on the MTS-Dialog dataset, in which conversations are short and usually do not contain information related to all SOAP categories.

\section{Experiments}

\subsection{Model details}
We used WangLab's FLAN-T5 and LED summarization models in the MEDIQA-Chat Challenge \footnote{\url{https://huggingface.co/wanglab/task-a-flan-t5-large-run-2}}\footnote{\url{https://huggingface.co/wanglab/task-b-led-large-16384-pubmed-run-3}}. To evaluate the FLAN-T5 model on input longer than its training data, we modify the maximum token length from 1024 to 4096. Table~\ref{tab:model_prompts} shows the prompts for all models in the experiments. The prompts of FLAN-T5 and LED follow WangLab's settings. For GPT models, we followed LED and FLAN-T5 prompts but removed the "including family history, diagnosis, past medical (and surgical) history, and known allergies" to prevent GPTs from specifically clarifying that certain information is not part of the conversation. Lastly, we designed a prompt to guide GPT in generating a summary with SOAP sections and a more parsable format.
 
\begin{table}[htbp!]
\centering
\includegraphics[scale=0.95]{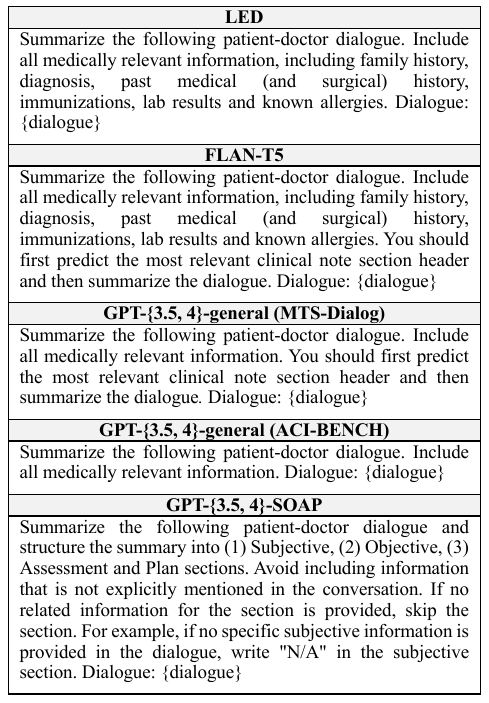}
\caption{Model prompts.}
\label{tab:model_prompts}
\end{table}

\subsection{Evaluation metrics}

All models were evaluated using ROUGE-1~\cite{lin2004rouge} and the average of ROUGE-1, BLEURT~\cite{sellam2020bleurt}, and BERTScore~\cite{zhang2019bertscore} (referred to as an aggregate score). These automatic metrics have been shown to correlate highly with human judgments for the doctor-patient conversations in recent studies~\cite{abacha2023investigation}. The section headers in the reference and generated notes were excluded from the evaluation. We used the \emph{en\_core\_sci\_sm} model in scispacy\footnote{\url{https://allenai.github.io/scispacy/}} to identify the medical terms in the dialogue and notes. Lastly, Linguistic Inquiry and Word Count (LIWC)~\cite{ tausczik2010psychological} was used to analyze the word distribution in SOAP notes. LIWC is a text analysis tool that systematically examines and categorizes language based on psychologically meaningful dimensions. It aids in deciphering the linguistic characteristics of written or spoken text, providing insights into the emotional and cognitive dimensions of communication. Because emotional and cognitive words can reflect aspects of a person's health in certain situations, they play essential roles in the SOAP note. 

\section{Results}

\subsection{Cross-dataset Performance}\label{sec:cross_dataset}

We evaluated the cross-dataset performance of doctor-patient conversation summarization models. Performance on the ACI-BENCH dataset is presented in Table~\ref{tab:cross_aci_rogue}. The experimental results indicate a notable performance decrease in out-of-domain models compared to the in-domain baseline (i.e., LED). We also noticed that the general model performed particularly poorly on objective notes. When utilizing the general model for doctor-patient summarization, adaptations are essential to preserve objective information. A potential approach involves treating the generation of objective notes as a distinct task. For example, the outcomes from gpt-SOAP models indicate that the performance of objective notes increases greatly by specifically instructing the model to generate notes with an objective section.

\begin{table}[hptb!]
\centering
\resizebox{\columnwidth}{!}{%
\begin{tabular}{|cccc|}
\hline
\multicolumn{1}{|c|}{\diagbox{Model}{Testing data}}                                                               & \multicolumn{1}{c|}{S}                                                    & \multicolumn{1}{c|}{O}                                                    & AP                           \\ \hline \hline
\multicolumn{4}{|c|}{ROUGE-1}                                                                                                                                                                                                                                               \\ \hline
\rowcolor[HTML]{EFEFEF} 
\multicolumn{1}{|c|}{\cellcolor[HTML]{EFEFEF}{\color[HTML]{000000} LED (In-domain)}} & \multicolumn{1}{c|}{\cellcolor[HTML]{EFEFEF}{\color[HTML]{000000} 0.554}} & \multicolumn{1}{c|}{\cellcolor[HTML]{EFEFEF}{\color[HTML]{000000} 0.502}} & {\color[HTML]{000000} 0.491} \\ \hline
\multicolumn{1}{|c|}{gpt3.5-SOAP}                                                    & \multicolumn{1}{c|}{0.358 (-35\%)}                                        & \multicolumn{1}{c|}{0.420 (-16\%)}                                        & 0.381 (-22\%)                \\ \hline
\multicolumn{1}{|c|}{gpt4-SOAP}                                                      & \multicolumn{1}{c|}{0.373 (-33\%)}                                        & \multicolumn{1}{c|}{0.447 (-11\%)}                                        & 0.379 (-23\%)                \\ \hline \hline
\multicolumn{1}{|c|}{FLAN-T5}                                                        & \multicolumn{1}{c|}{0.339 (-39\%)}                                        & \multicolumn{1}{c|}{0.146 (-71\%)}                                        & 0.265 (-46\%)                \\ \hline
\multicolumn{1}{|c|}{gpt3.5-general}                                                 & \multicolumn{1}{c|}{0.349 (-37\%)}                                        & \multicolumn{1}{c|}{0.175 (-65\%)}                                        & 0.352 (-28\%)                \\ \hline
\multicolumn{1}{|c|}{gpt4-general}                                                   & \multicolumn{1}{c|}{0.370 (-33\%)}                                        & \multicolumn{1}{c|}{0.179 (-64\%)}                                        & 0.363 (-26\%)                \\ \hline \hline
\multicolumn{4}{|c|}{Aggregate score}                                                                                                                                                                                                                                       \\ \hline
\rowcolor[HTML]{EFEFEF} 
\multicolumn{1}{|c|}{\cellcolor[HTML]{EFEFEF}LED (In-domain)}                        & \multicolumn{1}{c|}{\cellcolor[HTML]{EFEFEF}0.569}                        & \multicolumn{1}{c|}{\cellcolor[HTML]{EFEFEF}0.538}                        & 0.546                        \\ \hline
\multicolumn{1}{|c|}{gpt3.5-SOAP}                                                    & \multicolumn{1}{c|}{0.494 (-13\%)}                                        & \multicolumn{1}{c|}{0.527 (-2\%)}                                         & 0.520 (-5\%)                 \\ \hline
\multicolumn{1}{|c|}{gpt4-SOAP}                                                      & \multicolumn{1}{c|}{0.504 (-11\%)}                                        & \multicolumn{1}{c|}{0.552 (+2\%)}                                         & 0.518 (-5\%)                 \\ \hline \hline
\multicolumn{1}{|c|}{FLAN-T5}                                                        & \multicolumn{1}{c|}{0.447 (-21\%)}                                        & \multicolumn{1}{c|}{0.350 (-35\%)}                                        & 0.407 (-25\%)                \\ \hline
\multicolumn{1}{|c|}{gpt3.5-general}                                                 & \multicolumn{1}{c|}{0.478 (-16\%)}                                        & \multicolumn{1}{c|}{0.384 (-29\%)}                                        & 0.479 (-12\%)                \\ \hline
\multicolumn{1}{|c|}{gpt4-general}                                                   & \multicolumn{1}{c|}{0.487 (-14\%)}                                        & \multicolumn{1}{c|}{0.395 (-27\%)}                                        & 0.482 (-12\%)                \\ \hline
\end{tabular}
}
\caption{Model performance on the ACI-BENCH dataset. Testing data S, O, and AP means the evaluated reference note is the subjective, objective, and assessment and plan sections of the original reference note, respectively. The values in parentheses indicate the performance change compared with in-domain LED model (i.e., LED fine-tuned on ACI-BENCH). The FLAN-T5 model is fine-tuned on the MTS-Dialog dataset.}
\label{tab:cross_aci_rogue}
\end{table}

Table~\ref{tab:cross_mts_rogue} shows performance on the MTS-Dialog dataset. Because the reference in the MTS-Dialog dataset only focuses on one category, we ignore unmatched sections of the generated note. For example, if the reference note has a subjective section header, we only compared the reference with the subjective section of the generated note (i.e., LED-S, gpt-3.5-SOAP-S, and gpt-4-SOAP-S). Results again reveal a notable performance decrease in out-of-domain models compared to the in-domain baseline (i.e., FLAN-T5). In addition, the performance of objective notes exhibits a relatively milder decline for the SOAP-oriented model.

\textbf{Finding 1 (RQ1)}: despite the high performance on the in-domain testing data, the fine-tuning LM-based summarization method suffers from overfitting issues, leading to a notable performance drop on out-of-domain data.

\textbf{Finding 2 (RQ2)}: When employing the general-purpose model for doctor-patient summarization, adaptation is essential to ensure the preservation of objective information, which is more prone to being excluded. Experimental results of gpt-SOAP models indicate that the performance of objective notes can be greatly improved by specifically instructing GPT to generate notes with an objective section.

\begin{table}[h]
\centering
\resizebox{\columnwidth}{!}{%
\begin{tabular}{|cccc|}
\hline
\multicolumn{1}{|c|}{\diagbox{Model}{Testing data}}                                            & \multicolumn{1}{c|}{S}                                     & \multicolumn{1}{c|}{O}                                     & AP            \\ \hline \hline
\multicolumn{4}{|c|}{ROUGE-1}            \\ \hline
\rowcolor[HTML]{EFEFEF} 
\multicolumn{1}{|c|}{\cellcolor[HTML]{EFEFEF}FLAN-T5 (In-domain)} & \multicolumn{1}{c|}{\cellcolor[HTML]{EFEFEF}0.449}         & \multicolumn{1}{c|}{\cellcolor[HTML]{EFEFEF}0.435}         & 0.405         \\ \hline
\multicolumn{1}{|c|}{gpt-3.5-general}                             & \multicolumn{1}{c|}{0.244 (-46\%)}                         & \multicolumn{1}{c|}{0.266 (-39\%)}                         & 0.180 (-55\%) \\ \hline
\multicolumn{1}{|c|}{gpt4-general}                                & \multicolumn{1}{c|}{0.315 (-30\%)}                         & \multicolumn{1}{c|}{0.298 (-31\%)}                         & 0.214 (-47\%) \\ \hline \hline
\multicolumn{1}{|c|}{LED-S}                                       & \multicolumn{1}{c|}{0.231 (-49\%)}                         & \multicolumn{1}{c|}{-}                                     & -             \\ \hline
\multicolumn{1}{|c|}{LED-O}                                       & \multicolumn{1}{c|}{-}                                     & \multicolumn{1}{c|}{0.259 (-40\%)}                         & -             \\ \hline
\multicolumn{1}{|c|}{LED-AP}                                       & \multicolumn{1}{c|}{-}                                     & \multicolumn{1}{c|}{-}                                     & 0.112 (-72\%) \\ \hline
\rowcolor[HTML]{EFEFEF} 
\multicolumn{1}{|c|}{\cellcolor[HTML]{EFEFEF}gpt-3.5-SOAP-S}      & \multicolumn{1}{c|}{\cellcolor[HTML]{EFEFEF}0.225 (-50\%)} & \multicolumn{1}{c|}{\cellcolor[HTML]{EFEFEF}-}             & -             \\ \hline
\rowcolor[HTML]{EFEFEF} 
\multicolumn{1}{|c|}{\cellcolor[HTML]{EFEFEF}gpt-3.5-SOAP-O}      & \multicolumn{1}{c|}{\cellcolor[HTML]{EFEFEF}-}             & \multicolumn{1}{c|}{\cellcolor[HTML]{EFEFEF}0.357 (-18\%)} & -             \\ \hline
\rowcolor[HTML]{EFEFEF} 
\multicolumn{1}{|c|}{\cellcolor[HTML]{EFEFEF}gpt-3.5-SOAP-AP}     & \multicolumn{1}{c|}{\cellcolor[HTML]{EFEFEF}-}             & \multicolumn{1}{c|}{\cellcolor[HTML]{EFEFEF}-}             & 0.143 (-65\%) \\ \hline
\multicolumn{1}{|c|}{gpt-4-SOAP-S}                                & \multicolumn{1}{c|}{0.273 (-39\%)}                         & \multicolumn{1}{c|}{-}                                     & -             \\ \hline
\multicolumn{1}{|c|}{gpt-4-SOAP-O}                                & \multicolumn{1}{c|}{-}                                     & \multicolumn{1}{c|}{0.347 (-20\%)}                         & -             \\ \hline
\multicolumn{1}{|c|}{gpt-4-SOAP-AP}                               & \multicolumn{1}{c|}{-}                                     & \multicolumn{1}{c|}{-}                                     & 0.184 (-55\%) \\ \hline \hline
\multicolumn{4}{|c|}{Aggregate Score}                                                                                                                                                                       \\ \hline
\rowcolor[HTML]{EFEFEF} 
\multicolumn{1}{|c|}{\cellcolor[HTML]{EFEFEF}FLAN-T5 (In-domain)} & \multicolumn{1}{c|}{\cellcolor[HTML]{EFEFEF}0.584}         & \multicolumn{1}{c|}{\cellcolor[HTML]{EFEFEF}0.540}         & 0.545         \\ \hline
\multicolumn{1}{|c|}{gpt-3.5-general}                             & \multicolumn{1}{c|}{0.460 (-21\%)}                         & \multicolumn{1}{c|}{0.465 (-14\%)}                         & 0.423 (-22\%) \\ \hline
\multicolumn{1}{|c|}{gpt4-general}                                & \multicolumn{1}{c|}{0.513 (-12\%)}                         & \multicolumn{1}{c|}{0.480 (-11\%)}                         & 0.449 (-18\%) \\ \hline \hline
\multicolumn{1}{|c|}{LED-S}                                       & \multicolumn{1}{c|}{0.401 (-31\%)}                         & \multicolumn{1}{c|}{-}                                     & -             \\ \hline
\multicolumn{1}{|c|}{LED-O}                                       & \multicolumn{1}{c|}{-}                                     & \multicolumn{1}{c|}{0.411 (-24\%)}                         & -             \\ \hline
\multicolumn{1}{|c|}{LED-AP}                                       & \multicolumn{1}{c|}{-}                                     & \multicolumn{1}{c|}{-}                                     & 0.334 (-39\%) \\ \hline
\rowcolor[HTML]{EFEFEF} 
\multicolumn{1}{|c|}{\cellcolor[HTML]{EFEFEF}gpt-3.5-SOAP-S}      & \multicolumn{1}{c|}{\cellcolor[HTML]{EFEFEF}0.408 (-30\%)} & \multicolumn{1}{c|}{\cellcolor[HTML]{EFEFEF}-}             & -             \\ \hline
\rowcolor[HTML]{EFEFEF} 
\multicolumn{1}{|c|}{\cellcolor[HTML]{EFEFEF}gpt-3.5-SOAP-O}      & \multicolumn{1}{c|}{\cellcolor[HTML]{EFEFEF}-}             & \multicolumn{1}{c|}{\cellcolor[HTML]{EFEFEF}0.482 (-11\%)} & -             \\ \hline
\rowcolor[HTML]{EFEFEF} 
\multicolumn{1}{|c|}{\cellcolor[HTML]{EFEFEF}gpt-3.5-SOAP-AP}     & \multicolumn{1}{c|}{\cellcolor[HTML]{EFEFEF}-}             & \multicolumn{1}{c|}{\cellcolor[HTML]{EFEFEF}-}             & 0.310 (-43\%) \\ \hline
\multicolumn{1}{|c|}{gpt-4-SOAP-S}                                & \multicolumn{1}{c|}{0.466 (-20\%)}                         & \multicolumn{1}{c|}{-}                                     & -             \\ \hline
\multicolumn{1}{|c|}{gpt-4-SOAP-O}                                & \multicolumn{1}{c|}{-}                                     & \multicolumn{1}{c|}{0.492 (-9\%)}                          & -             \\ \hline
\multicolumn{1}{|c|}{gpt-4-SOAP-AP}                               & \multicolumn{1}{c|}{-}                                     & \multicolumn{1}{c|}{-}                                     & 0.406 (-26\%) \\ \hline
\end{tabular}
}
\caption{Model performance on the MTS-Dialog dataset. Testing data S, O, and AP means that the evaluated reference note belongs to the subjective, objective, and assessment and plan categories, respectively. \emph{-S}, \emph{-O}, and \emph{-AP} indicate the generated note in the subjective, objective, and assessment and plan sections, respectively. The values in parentheses indicate the performance change compared with the in-domain FLAN-T5 model (i.e., FLAN-T5 model fine-tuned on MTS-Dialog).}
\label{tab:cross_mts_rogue}
\end{table}

\subsection{LIWC Analysis of SOAP Note}

Experimental results presented in Section~\ref{sec:cross_dataset} reveal a notable decline in the performance of the fine-tuning language model-based method when applied to out-of-domain data. In this section, we investigate the characteristics of S, O, and AP samples in two datasets to better understand potential factors for performance degradation.

\begin{figure*}
\centering
\includegraphics[scale=0.9]{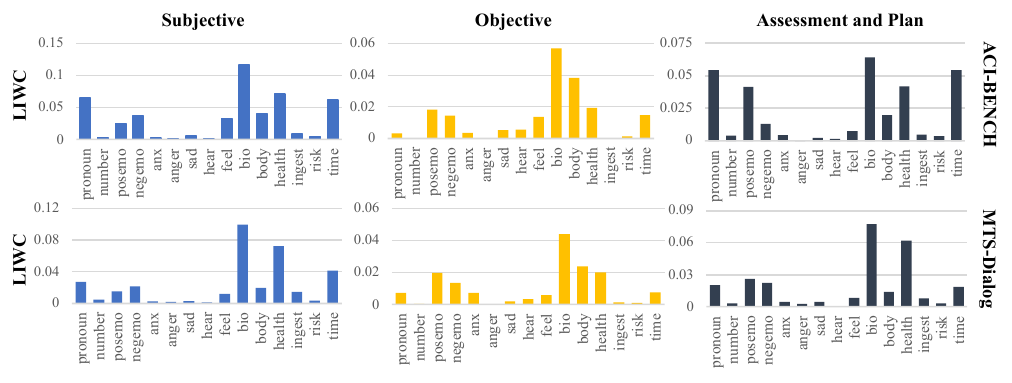}
\caption{LIWC analysis of SOAP notes. Note that this result is calculated using all samples (i.e., training, validation, and testing sets), rather than using only the testing set as experiments on model performance. In addition, for simplicity and visualization purpose, we only show that LIWC categories that have a higher association with healthcare. The correlations between the two data sets are 0.93, 0.95, and 0.77 in S, O, and AP, respectively.
}
\label{fig:LIWC_analysis}
\end{figure*}

We computed LIWC features for S, O, and AP notes. Table~\ref{table:LIWC_example} shows the example words in the selected LIWC categories, and Figure~\ref{fig:LIWC_analysis} visualizes the selected LIWC features for the ACI-BENCH and MTS-Dialog datasets. First, we find that LIWC shares similar patterns for S, O, and AP notes across the ACI-BENCH and MTS-Dialog datasets. Specifically, these datasets have corrections of 0.93, 0.95, and 0.77 for S, O, and AP notes, respectively. These results indicate that the SOAP notes in the two datasets are structured in a similar way in terms of word category distribution. 

We also observe a similarity in LIWC features between S and AP notes. This alignment is intuitive as S represents subjective information provided by the patient, whereas AP represents the \emph{subjective} assessment and plan from the doctor. One difference between the S and AP notes is that, in S notes, negative emotion is higher than positive emotion, while in the AP notes, negative emotion is lower than positive emotion. This fits a typical scenario where a patient comes to the doctor because of concerns (negative emotion), and then the doctor makes an assessment and plans to address the patient’s problem, introducing a more positive emotion. 

\textbf{Finding 3}: LIWC features have characteristics that resonate with SOAP notes in real-world scenarios. 

\textbf{Finding 4 (RQ1)}: Because LIWC features exhibit strong correlations for S, O, and AP notes across different datasets, format mismatch (i.e., discrepancies in word distribution) might not be the main cause of the model's performance decline on out-of-domain data.

\begin{table}[htbp!]
\centering
\includegraphics[scale=0.8]{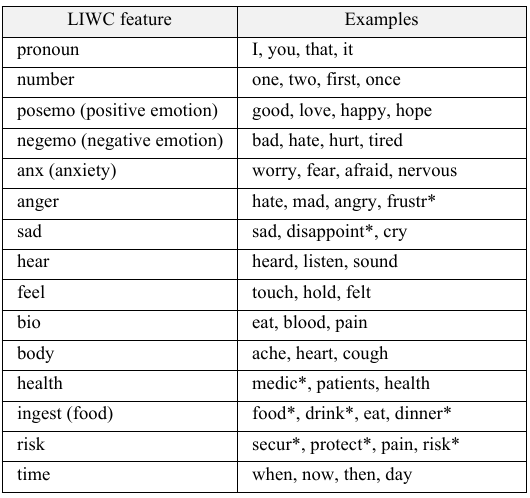}
\caption{Selected LIWC features and example words.}
\label{table:LIWC_example}
\end{table}

\subsection{Hallucination analysis}\label{sec:medical_term_exp}

\begin{figure}[ht]
\centering
\includegraphics[scale=0.85]{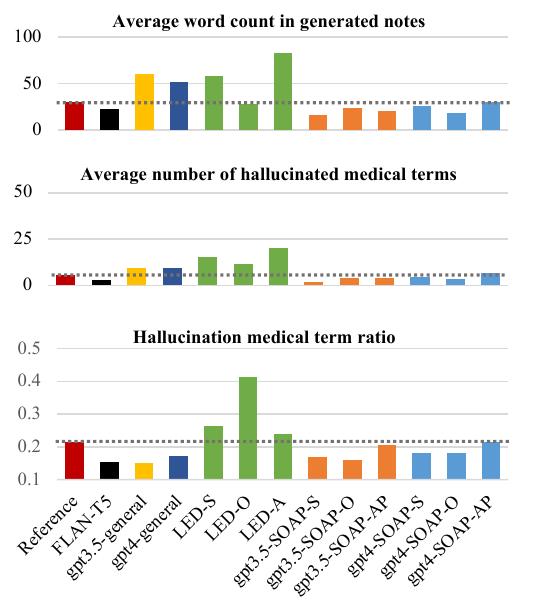}
\caption{Hallucination medical term ratio, the experiments were conducted on the MTS-Dialog dataset.}
\label{fig:hall_med_term}
\end{figure}

We examine the hallucination problem of SOAP-oriented models in scenarios where the input conversation might not include all SOAP information (Figure~\ref{fig:hall_med_term}.) First, we compute the length of the generated note. Because Flan-T5 is fine-tuned with the in-domain data, the resulting note lengths are closer to the reference than other models. In contrast, the out-of-domain LED model generated notes much longer than the reference. In the case of the SOAP-oriented GPT models, each section (S, O, and AP) is shorter than the general model, but the combination of all sections (gpt-S + gpt-O + gpt-AP) is slightly longer than that of the general GPT model. 

We then counted the number of unique medical terms that were not mentioned in the input dialogue but \textbf{\textit {were}} generated in the note (i.e., hallucinated medical terms). Finally, we divided the number of hallucinated medical terms by the length of the generated note to derive the hallucination ratio. We observed that LED has a notable hallucination problem for medical information especially on the objective section. For the SOAP-oriented GPT models, the AP sections (i.e., gpt3.5-SOAP-AP and gpt4-SOAP-AP) exhibit a relatively higher hallucination ratio, suggesting a higher tendency of generating hallucinated medical terms for assessment and plan section. Overall, the GPT-based models manifest a considerably lower hallucination ratio than LED in general. 

Even the reference note may contain medical terms not present in the original dialogue. Reasons for this discrepancy are that the reference note is abstractive summarization and may use synonyms as substitutes for the original words. For example, the word "flu" in the conversation was replaced with "influenza" in the reference note. However, we believe it would be better to use  exactly the same words as those in the conversation. Although LMs specialized in the medical domain might be aware of the similarity of medical terms, using the same medical terms as the conversation provide better consistency and can avoid confusion. In addition, as new medical terms emerge, the language model might not be updated with the latest information.

\subsection{Case study of SOAP-oriented model}

\begin{table*}
\centering
\includegraphics[scale=0.76]{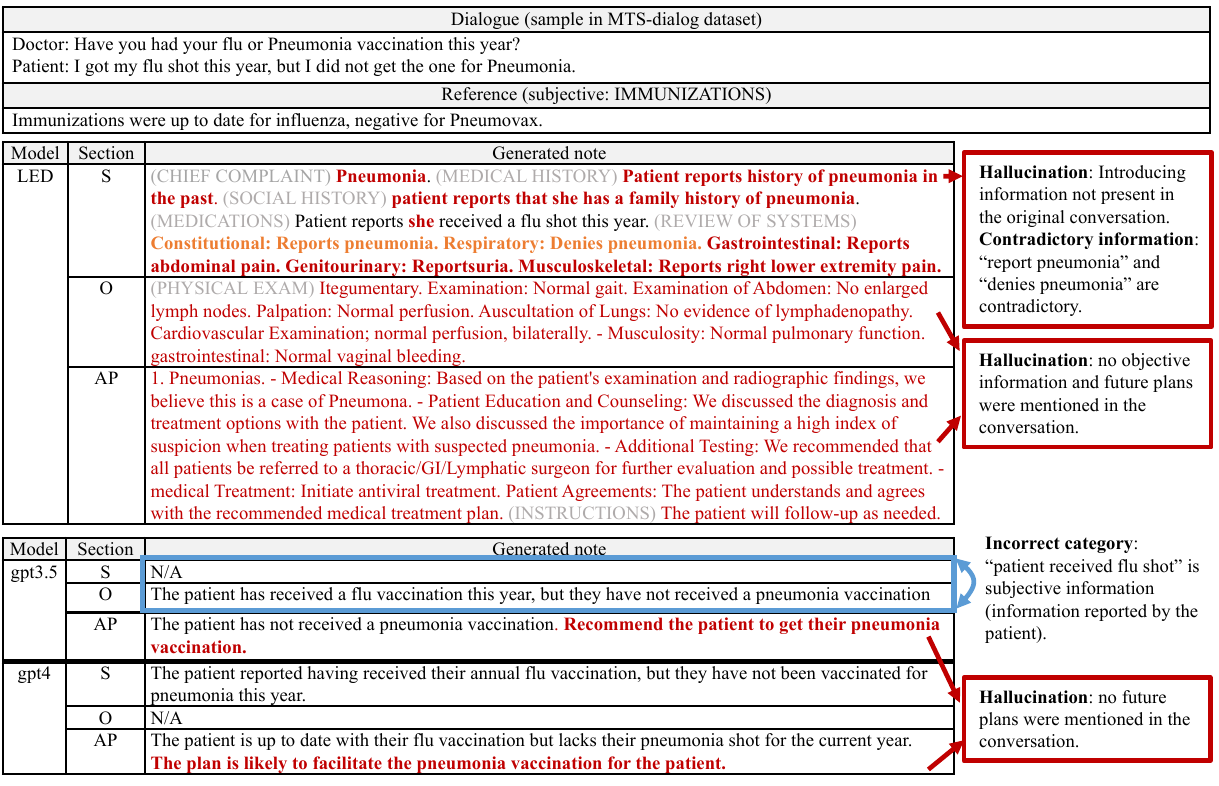}
\caption{Case study example}
\label{tab:case_study}
\end{table*}

We conducted a case study to better understand the errors made by the summarization models (Table~\ref{tab:case_study}). First, we observed that the LED model tends to overfit its training data. In cases where the input conversation lacks sufficient information, the LED model generates unfaithful and irrelevant information, trying to align the generated note more closely with its training data. In contrast, the gpt3.5-SOAP model has difficulty accurately classifying information. For example, "patient received flu shot" is information obtained from the patient and thus should be categorized as subjective information, but the gpt3.5 model incorrectly categorized this information as an objective fact. Moreover, although the prompts are instructed to “avoid including information that is not explicitly mentioned in the conversation (Table~\ref{tab:model_prompts})”, both gpt3.5-SOAP and gpt4-SOAP models produce hallucination results in the generated AP note. This aligns with our observation in Figure~\ref{fig:hall_med_term} that SOAP-oriented GPT models have a higher hallucination medical term ratio in the AP section. This result suggests that it is important to examine the assessment and plan section, as the model may have a higher tendency to generate hallucinated information in this category.

\section{Limitations}

One limitation of this study is that the SOAP data in the MTS-Dialog dataset is unbalanced, with most references focusing on subjective information. In addition, real-world doctor-patient conversations are complex in size and medical specialties and cannot be fully represented by two datasets. Another issue lies in the generative model producing varied results in different runs, and the performance of the GPT model is affected by the prompt.

\section{Conclusion}

In this study, we evaluated the SOTA doctor-patient summarization models on out-of-domain data and investigated the challenges of using fine-tuning LM and GPT-based summarization models in real-world applications. For a model with a general configuration, the results indicate a high tendency of omitting objective information in the generated note. This concern can be alleviated by adopting the SOAP-oriented configuration, which orients the model to generate information relevant to all essential categories. Despite achieving the highest performance on in-domain data, the fine-tuned LM with SOAP-oriented configuration exhibits a significant hallucination issue. To generate a note closer to its training data, the model produces hallucinations when none or insufficiently related information is present in the conversation. In contrast, limitations of GPT-based models arise from a tendency to offer their own suggestions for the assessment and plan. We hope our results provide insights for future work toward creating more robust models for real-world settings. 


\bibliography{anthology,custom}
\bibliographystyle{acl_natbib}

\end{document}